%% file: iswc23.tex
\documentclass[runningheads]{llncs}
\usepackage[T1]{fontenc}
\usepackage{graphicx}

\usepackage{amsmath}
\usepackage{tabularx}
\usepackage{booktabs}
\usepackage{makecell}
\usepackage{enumitem}
\usepackage[dvipsnames]{xcolor}
\usepackage{multirow}
\usepackage{amssymb}
\usepackage{pifont}

\newcommand{\cmark}{\textcolor{green}{\ding{51}}}
\newcommand{\xmark}{\textcolor{red}{\ding{55}}}
\newcommand{\pubgraph}{PubGraph}

\usepackage[colorlinks = true,
            linkcolor = blue,
            urlcolor  = blue,
            citecolor = blue,
            anchorcolor = blue]{hyperref}
            
\usepackage{color}

\begin{document}
\title{PubGraph: A Large-Scale Scientific Knowledge Graph}
%
%
\author{
    Kian Ahrabian \and
    Xinwei Du \and
    Richard Delwin Myloth \and
    Arun Baalaaji Sankar Ananthan \and
    Jay Pujara
}
\authorrunning{K. Ahrabian et al.}
%
\institute{University of Southern California, Information Sciences Institute, Marina del Rey CA 90292, USA \\
\email{\{ahrabian,xinweidu,myloth,arunbaal,jpujara\}@usc.edu}}


%
\maketitle              

\input{sections/abstract}
\input{sections/intro}
\input{sections/dataset}
\input{sections/related}
\input{sections/conclusion}

\section*{Acknowledgements}
This work was funded by the Defense Advanced Research Projects Agency with award W911NF-19-20271 and with support from a Keston Exploratory Research Award.

\paragraph*{Resource Availability Statement:}
The source code for building PubGraph, along with a data schema, is available from GitHub, released under the CC-BY-SA license\footnote{https://github.com/usc-isi-i2/isi-pubgraph}.
All the introduced benchmarks and resources are publicly accessible and released under the CC-BY-SA license\footnote{https://pubgraph.isi.edu/}.
Due to the sheer size of the resources (> 2TB), we could not host the data in any commonly used platform and had to resort to self-provisioned servers.

\bibliographystyle{splncs04}
\bibliography{main}



\end{document}

%% file: sections/abstract.tex
\begin{abstract}
Research publications are the primary vehicle for sharing scientific progress in the form of new discoveries, methods, techniques, and insights. 
Unfortunately, the lack of a large-scale, comprehensive, and easy-to-use resource capturing the myriad relationships between publications, their authors, and venues presents a barrier to applications for gaining a deeper understanding of science.
In this paper, we present \textbf{PubGraph}, a new resource for studying scientific progress that takes the form of a large-scale knowledge graph (KG) with more than 385M entities, 13B main edges, and 1.5B qualifier edges.
\pubgraph~is comprehensive and unifies data from various sources, including Wikidata, OpenAlex, and Semantic Scholar, using the Wikidata ontology.
Beyond the metadata available from these sources, PubGraph includes outputs from auxiliary community detection algorithms and large language models.
To further support studies on reasoning over scientific networks, we create several large-scale benchmarks extracted from PubGraph for the core task of knowledge graph completion (KGC).
These benchmarks present many challenges for knowledge graph embedding models, including an adversarial community-based KGC evaluation setting, zero-shot inductive learning, and large-scale learning.
All of the aforementioned resources are accessible at \href{https://pubgraph.isi.edu/}{https://pubgraph.isi.edu/} and released under the CC-BY-SA license.
We plan to update PubGraph quarterly to accommodate the release of new publications.

\keywords{Scientific Knowledge Graphs \and Knowledge Graph Completion \and Inductive Learning}

\end{abstract}

%% file: sections/intro.tex
\section{Introduction}

Scientific progress takes many forms, from discovering new species to repurposing extant models for novel tasks. 
Innovation in science has been studied from a variety of perspectives, including the combination of scholarly domains~\cite{hofstra2020diversity,uzzi2013atypical}, sociological factors~\cite{de2015game}, and analogical reasoning~\cite{hope2017accelerating,kang2022augmenting}.
However, many studies of this phenomenon have been limited due to the difficulty in finding and using large-scale data for the domain.
In this paper, we address this obstacle by introducing \textbf{PubGraph}, a knowledge graph (KG) with new resources and benchmarks, enabling the study of scientific research at scale using structural patterns in citation and collaboration networks.
PubGraph also provides a unique opportunity to compare models on core tasks such as transductive and inductive knowledge graph completion (KGC).

\begin{figure}[ht!]
  \centering
  \includegraphics[width=0.97\textwidth]{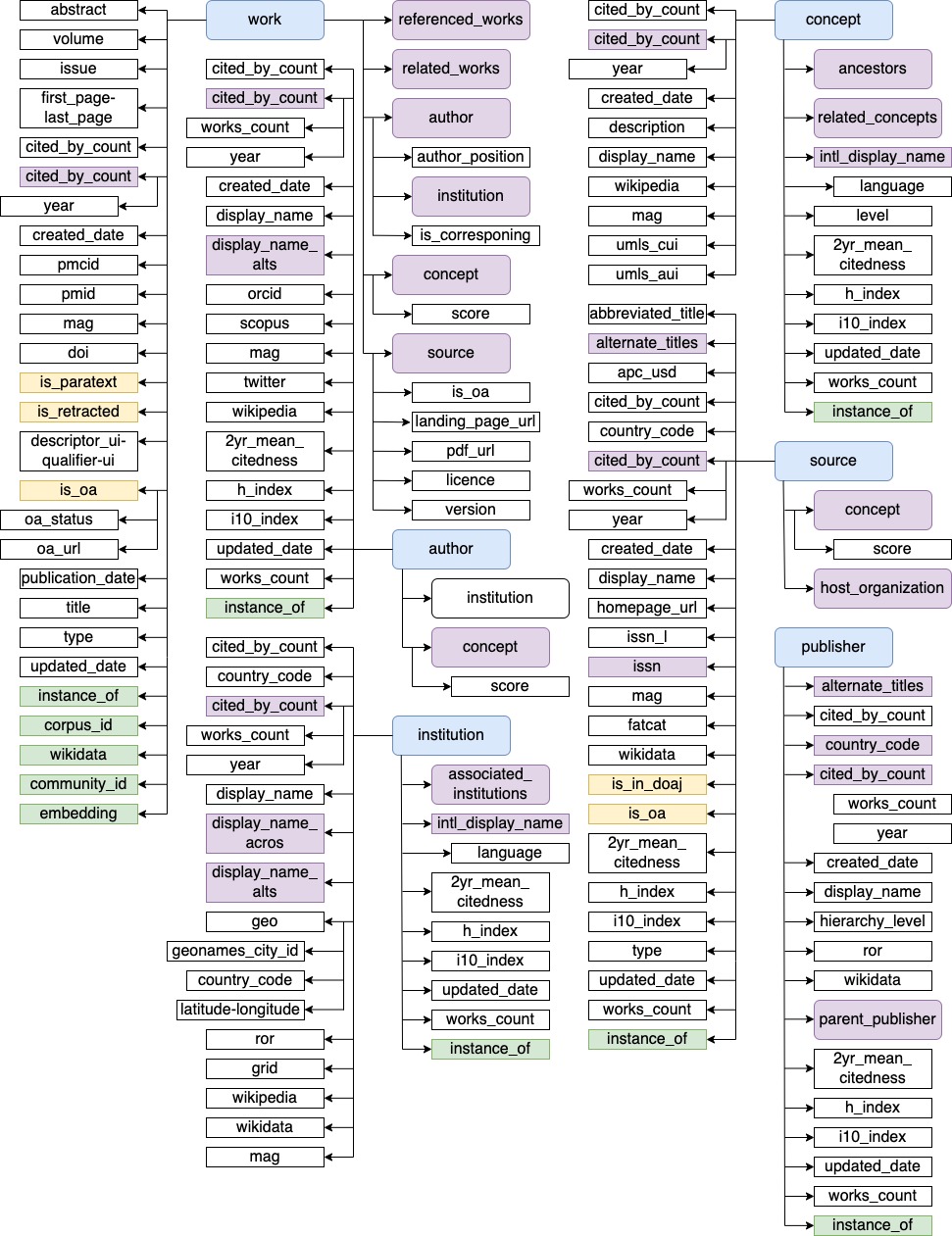}
  \caption{Overview of PubGraph schema. \textbf{Legend.} \textit{Colors:} Blue $\rightarrow$ Main entity, Yellow $\rightarrow$ Boolean attribute, Purple $\rightarrow$ Multi attribute, and Green $\rightarrow$ New attribute; \textit{Shapes:} Rounded rectangle $\rightarrow$ Entity attribute, and Rectangle $\rightarrow$ Regular attribute.}
  \label{fig:data:over}
\end{figure}

PubGraph is a large-scale multi-relational KG built on top of the OpenAlex catalog~\cite{priem2022openalex} and the Wikidata~\cite{vrandevcic2014wikidata} ontology.
It consists of more than 385M entities, comprising authors, institutions, sources, papers, and concepts, and more than 13B main edges and 1.5B qualifier edges among those entities.
PubGraph captures temporal information, allowing the study of scientific works' dynamics.
Additionally, it also connects the scholarly articles available in OpenAlex to their counterparts in the Semantic Scholar Academic Graph (S2AG)~\cite{wade2022semantic} and Wikidata through external ids.
Moreover, besides the metadata information available in OpenAlex, PubGraph provides outputs from auxiliary community detection algorithms and large language models to further assist future studies of scientific articles.
Fig. \ref{fig:data:over} illustrates an overview of PubGraph schema.
In this paper, we describe the methodology used to construct PubGraph, i.e., the ontological choices made for mapping OpenAlex to Wikidata, the model choices to extract outputs from auxiliary models, and the entity resolution procedure for mapping OpenAlex articles to S2AG and Wikidata.

One of the essential parts of studying scientific progress is understanding and reasoning about connections between ideas and discoveries.
However, there is a shortage of benchmarks that could be used to study such topics.
In the past, citations have proven to be crucial in studying publications and their impact~\cite{price1965networks}.
Prior works have also studied tasks on citations such as intent classification~\cite{cohan2019structural,gururangan2020don,jurgens2018measuring}, recommendation~\cite{bhagavatula2018content,farber2020hybridcite}, and prediction~\cite{cohan2020specter,ostendorff2022neighborhood}.
In this work, we introduce new large-scale benchmarks for finding connections among scientific works framed as a KGC task.
The KGC task requires models to predict a target entity, given a source entity and a relation.
The aim of this task is to support the study of citations from a structural perspective in both transductive, i.e., all nodes are known, and inductive, i.e., evaluation nodes are unseen, settings.
Moreover, we also identify a community-based adversarial evaluation setting that mitigates the influence of random negative sampling in the evaluation phase of large-scale KGs.

The contributions of this work are summarized as follows:
\begin{enumerate}[topsep=0pt]
  \item Introducing PubGraph, a billion-scale, multi-relational KG built on top of the OpenAlex catalog
  \item Mapping the OpenAlex metadata to Wikidata ontology
  \item Connecting two other large-scale scholarly metadata repositories, S2AG and Wikidata, to make PubGraph a unifying and comprehensive resource
  \item Introducing large-scale extrapolated KGC benchmarks for KG models in both transductive and inductive settings
  \item Identifying challenging adversarial evaluation settings for KGC benchmarks
\end{enumerate}

%% file: sections/dataset.tex
\section{Building PubGraph}
The primary source for creating PubGraph is the metadata in the OpenAlex catalog that we map to the Wikidata ontology.
OpenAlex is an open-source catalog of scholarly entities that provides metadata for works, authors, institutions, sources, publishers, and concepts.
Moreover, we add connections to both S2AG and Wikidata repositories to provide a more unifying resource for the researchers.
Furthermore, we provide outputs from auxiliary models to further enrich PubGraph for future studies.
The rest of this section is organized as follows: Sec. \ref{pubg:ont} introduces the mapping procedure from OpenAlex metadata to Wikidata ontology, Sec. \ref{pubg:res} describes the implemented procedure to connect S2AG and Wikidata with OpenAlex along with some statistics of the resolution, and Sec. \ref{pubg:aux} presents the model choices for auxiliary outputs included in PubGraph.


\subsection{Mapping to Wikidata Ontology}
\label{pubg:ont}
To transform the OpenAlex dump (taken on April 9th, 2023) into PubGraph, we follow the well-known and well-studied Wikidata ontology.
Specifically, we create a mapping between metadata information from the OpenAlex dump to Wikidata properties.
Using Wikidata enables broader adoption of the KG and clear semantics for entities and relationships.

\begin{table*}[t!]
    \caption{OpenAlex metadata mapping to properties covered by Wikidata ontology.}
    \label{tab:app:ont}
    \centering
    \resizebox{\textwidth}{!}{\begin{tabular}{c|c||c|c}
        \toprule
        \textbf{\makecell{OpenAlex Metadata}} & \textbf{\makecell{WikiData Property}} & \textbf{\makecell{OpenAlex Metadata}} & \textbf{\makecell{WikiData Property}} \\ \midrule
        \midrule
        abstract & \makecell{P7535} & author & \makecell{P50} \\ \midrule
        author position & \makecell{P1545} & institution & \makecell{P1416} \\ \midrule
        landing page url & \makecell{P973} & pdf url & \makecell{P953} \\ \midrule
        license & \makecell{P275} & version & \makecell{P9767} \\ \midrule
        volume & \makecell{P478} & issue & \makecell{P433} \\ \midrule
        \makecell{first page + last page} & \makecell{P304} & concept & \makecell{P921} \\ \midrule
        score & \makecell{P4271} & year & \makecell{P585} \\ \midrule
        created date & \makecell{P571} & doi & \makecell{P356} \\ \midrule
        mag & \makecell{P6366} & pmid & \makecell{P698} \\ \midrule
        pmcid & \makecell{P932} & \makecell{descriptor ui + qualifier ui} & \makecell{P9340} \\ \midrule
        oa status & \makecell{P6954} & oa url & \makecell{P2699} \\ \midrule
        publication date & \makecell{P577} & referenced work & \makecell{P2860} \\ \midrule
        title & \makecell{P1476} & type & \makecell{P31} \\ \midrule
        updated date & \makecell{P5017} & works count & \makecell{P3740} \\ \midrule
        display name & \makecell{P2561} & display name alternatives & \makecell{P4970} \\ \midrule
        orcid & \makecell{P496} & scopus & \makecell{P1153} \\ \midrule
        twitter & \makecell{P2002} & wikipedia & \makecell{P4656} \\ \midrule
        last known institution & \makecell{P1416} & abbreviated title & \makecell{P1813} \\ \midrule
        alternate titles & \makecell{P1476} & apc usd & \makecell{P2555} \\ \midrule
        country code & \makecell{P297} & homepage url & \makecell{P856} \\ \midrule
        host organization & \makecell{P749} & issn-l & \makecell{P7363} \\ \midrule
        issn & \makecell{P236} & fatcat & \makecell{P8608} \\ \midrule
        associated institution & \makecell{P1416} & relationship & \makecell{P1039} \\ \midrule
        display name acronyms & \makecell{P1813} & homepage url & \makecell{P856} \\ \midrule
        geonames city id & \makecell{P1566} & \makecell{latitude + longitude} & \makecell{P625} \\ \midrule
        ror & \makecell{P6782} & grid & \makecell{P2427} \\ \midrule
        international display name & \makecell{P4970} & language & \makecell{P9753} \\ \midrule
        level & \makecell{P1545} & alternate titles & \makecell{P4970} \\ \midrule
        hierarchy level & \makecell{P1545} & parent publisher & \makecell{P749} \\ \midrule
        location & \makecell{P1433} & ancestor & \makecell{P4900} \\ \midrule
        related concept & \makecell{P921} & corpus id & \makecell{P8299} \\ \midrule
        \end{tabular}}
\end{table*}

\begin{table*}[ht!]
    \caption{OpenAlex metadata mapping to properties not covered by Wikidata ontology.}
    \label{tab:app:art}
    \centering
    \resizebox{\textwidth}{!}{\begin{tabular}{c|c||c|c}
        \toprule
        \textbf{\makecell{OpenAlex Metadata}} & \textbf{\makecell{New Property}} & \textbf{\makecell{OpenAlex Metadata}} & \textbf{\makecell{New Property}} \\ \midrule
        \midrule
        best oa location & \makecell{P\_best\_oa\_location} & cited by count & \makecell{P\_total\_cited\_by\_count} \\ \midrule
        cited by count & \makecell{P\_cited\_by\_count} & primary location & \makecell{P\_primary\_location} \\ \midrule
        2yr mean citedness & \makecell{P\_impact\_factor} & h-index & \makecell{P\_h\_index} \\ \midrule
        i10-index & \makecell{P\_i10\_index} & wikidata & \makecell{P\_wikidata} \\ \midrule
        umls aui & \makecell{P\_umls\_aui} & community id & \makecell{P\_community\_id} \\ \midrule
        \end{tabular}}
\end{table*}

\begin{table*}[ht!]
    \caption{OpenAlex boolean metadata mapping to edges using Wikidata ontology.}
    \label{tab:app:bool}
    \centering
    \resizebox{\textwidth}{!}{\begin{tabular}{c|c||c|c}
        \toprule
        \textbf{\makecell{OpenAlex Metadata}} & \textbf{\makecell{Edge}} & \textbf{\makecell{OpenAlex Metadata}} & \textbf{\makecell{Edge}} \\ \midrule
        \midrule
        is corresponding & \makecell{P31 $\rightarrow$ Q36988860} & is oa & \makecell{P31 $\rightarrow$ Q232932} \\ \midrule
        is paratext & \makecell{P31 $\rightarrow$ Q853520} & is retracted & \makecell{P31 $\rightarrow$ Q45182324} \\ \midrule
        is in doaj & \makecell{P31 $\rightarrow$ Q1227538} & & \\ \midrule
        \end{tabular}}
\end{table*}

\begin{table*}[ht!]
    \caption{OpenAlex entity type mapping to edges using Wikidata ontology.}
    \label{tab:app:ent}
    \centering
    \resizebox{\textwidth}{!}{\begin{tabular}{c|c||c|c}
        \toprule
        \textbf{\makecell{OpenAlex Metadata}} & \textbf{\makecell{Edge}} & \textbf{\makecell{OpenAlex Metadata}} & \textbf{\makecell{Edge}} \\ \midrule
        \midrule
        work & \makecell{P31 $\rightarrow$ Q13442814} & author & \makecell{P31 $\rightarrow$ Q482980} \\ \midrule
        source & \makecell{P31 $\rightarrow$ Q1711593} & institution & \makecell{P31 $\rightarrow$ Q178706} \\ \midrule
        concept & \makecell{P31 $\rightarrow$ Q115949945} & publisher & \makecell{P31 $\rightarrow$ Q2085381} \\ \midrule
        \end{tabular}}
\end{table*}

Table \ref{tab:app:ont} presents the mapping from OpenAlex metadata to Wikidata properties.
These mappings are selected such that they best describe the metadata field.
Here, we explain the ontological design choices that we made for the mapping:
\begin{enumerate}
    \item abstract $\rightarrow$ P7535: Due to the absence of a one-to-one match, we use P7535 (scope and content), which is defined as ``a summary statement providing an overview of the archival collection.''
    \item author position $\rightarrow$ P1545: Since this field defines an order of the authors, we use P1545 (series ordinal), which is defined as the ``position of an item in its parent series (most frequently a 1-based index).''
    \item first page + last page $\rightarrow$ P304: Since OpenAlex uses two different fields to present this information, we merge them into one attribute to be aligned with the Wikidata ontology.
    \item score $\rightarrow$ P4271: Since this field indicates the relatedness of two concepts as produced by a model, it matches the definition of P4271 (rating) defined as ``qualifier to indicate a score given by the referenced source indicating the quality or completeness of the statement.''
    \item descriptor ui + qualifier ui $\rightarrow$ P1038: Since OpenAlex uses two different fields to present this information, we merge them into one attribute to be aligned with the Wikidata ontology.
    \item apc usd $\rightarrow$ P2555: Since this field describes a ``source's article processing charge in US Dollars'', we match it to P2555 (fee) defined as ``fee or toll payable to use, transit or enter the subject.''
    \item relationship $\rightarrow$ P1039: Since this field describes the relation between two institutions, we use P1039 (kinship to subject) defined as ``qualifier of "relative" (P1038) to indicate less usual family relationships.''
    \item location $\rightarrow$ P1433: Since this field describes the publishing location of a work, we match it with P1433 (published in).
    \item latitude + longitude $\rightarrow$ P625: Since OpenAlex uses two different fields to present this information, we merge them into one attribute to be aligned with the Wikidata ontology.
    \item level $\rightarrow$ P1545 and hierarchy level $\rightarrow$ P1545: Since there is no Wikidata property to describe a position in a hierarchy, we use the closest property P1545 (series ordinal), which is defined as the ``position of an item in its parent series (most frequently a 1-based index).''
\end{enumerate}

For the metadata with no suitable parallel property, we create new ones to keep the KG as complete as possible, as showcased in Table \ref{tab:app:art}.
Note that for ``cited by count'', OpenAlex provides both yearly and total values; hence, the reason for having two different new properties.
Moreover, for metadata with a boolean type, we add a new edge (main or qualifier) when true.
Table \ref{tab:app:bool} presents the edges representing each boolean metadata with all the relations and entities taken from the Wikidata repository.
This choice was made to maintain a better semantic composure and avoid creating new properties in the KG.
For example, there is no property in Wikidata for ``is paratext''; however, there exists an \textit{paratext} entity (Q853520).
Hence, instead of creating new property such as \textit{P\_is\_paratext}, we can create a new edge when ``is paratext'' is true to this entity with relation P31 (instance of).
Finally, we also add ``instance of'' edges to indicate the type of each entity as classified by OpenAlex, as presented in Table \ref{tab:app:ent}.
Given its flexibility to represent attributed graphs, we use $\text{RDF}^*$ as the graph representation for PubGraph (as illustrated in Fig. \ref{fig:data:over}).


\subsection{S2AG and WikiData Entity Resolution}
\label{pubg:res}
\begin{figure*}[t]
  \centering
  \includegraphics[width=\textwidth]{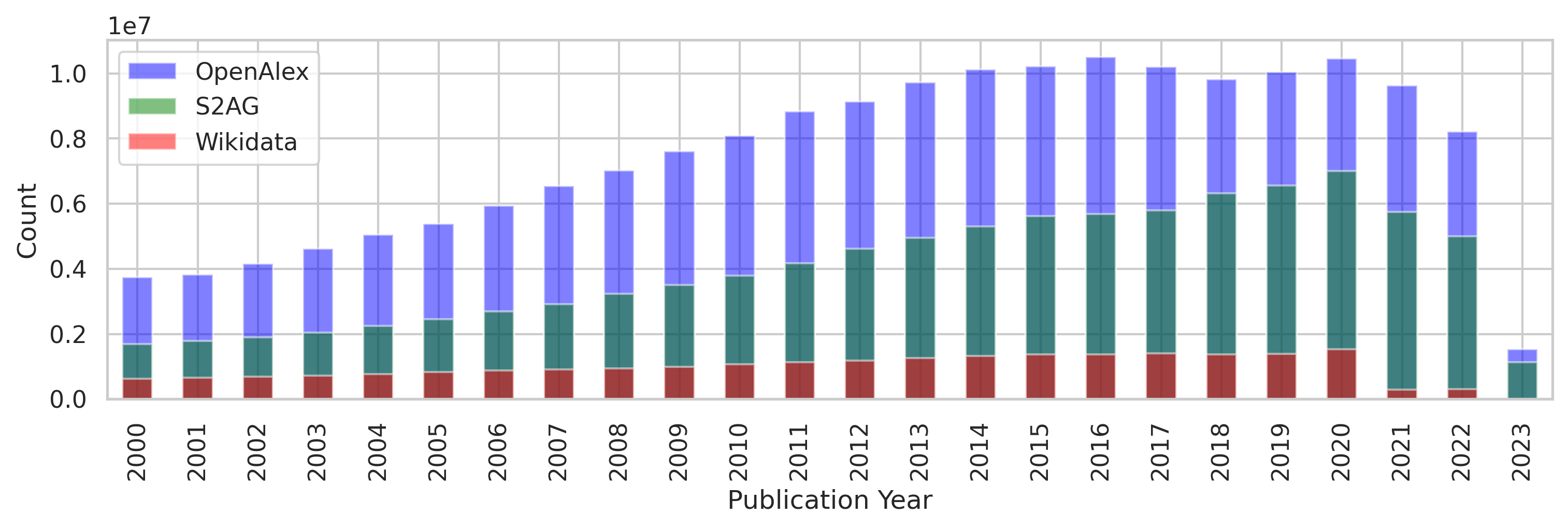}
  \caption{Distribution of publication years in the 2000-2023 period for OpenAlex, S2AG, and Wikidata. Note that only $\sim$128.3M out of the $\sim$211.5M papers in S2AG have publication dates and are included.}
  \label{fig:pubg:dist}
\end{figure*}

To make PubGraph a more unifying and comprehensive resource, we opt to connect works in OpenAlex to two other large-scale repositories of scholarly metadata: S2AG (taken on April 11th, 2023) and Wikidata (taken on April 28th, 2023).
Fig. \ref{fig:pubg:dist} showcases the distribution of publication years in the 2000-2023 period for the works available in these three repositories.
During this analysis, we noticed that only $\sim$128.3M out of the $\sim$211.5M papers in S2AG have publication dates.
This finding further highlights the importance of a unifying and comprehensive resource.
To this end, we follow a two-step procedure.
First, we match entities based on the following IDs: DOI, MAG, PMID, and PMCID.
For S2AG, this results in $\sim$197.6M out of $\sim$211.5M unique papers being matched to OpenAlex works, roughly providing a 93.4\% coverage.
For Wikidata, this results in $\sim$33.2M out of $\sim$38.9M unique papers being matched to OpenAlex works, roughly providing an 85.4\% coverage.

\begin{figure*}[t!]
  \centering
  \includegraphics[width=\textwidth]{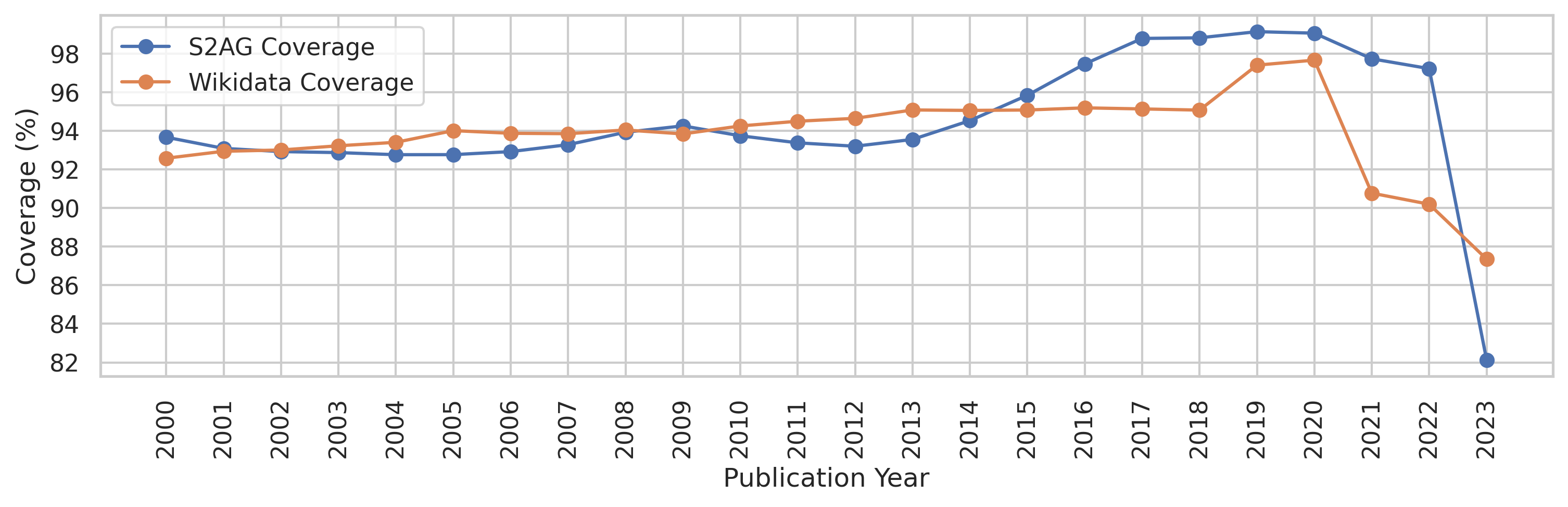}
  \caption{Coverage of S2AG and Wikidata papers after entity resolution in the 2000-2023 period.}
  \label{fig:pubg:cov}
\end{figure*}

Then, among the remaining unmatched entities, we run an exact title search and only keep one-to-one mappings.
For S2AG, this step further increases the number of matched unique papers to $\sim$199.2M, roughly providing a 94.2\% coverage.
For Wikidata, this step further increases the number of matched unique papers to $\sim$36.4M, roughly providing a 93.6\% coverage.
Fig. \ref{fig:pubg:cov} provides a coverage distribution over the 2000-2023 period for both S2AG and Wikidata.
As evident from this distribution, the coverage of both data sources seems to be relatively unbiased toward the time of publication.
We believe the Wikidata drop from 2021 onward is due to the low number of papers available in the platform in the period, and the S2AG drop is due to the potential delays in adding recent publications. 
Moreover, regarding more recent data, Wikidata seems to benefit drastically from adding new entities through external sources.
We plan to improve our entity resolution heuristic using other metadata, such as authors, to cover more entities in future releases.


\subsection{Auxiliary Outputs}
\label{pubg:aux}

\subsubsection{Community Detection}
Besides sharing scientific findings, scholarly articles represent the research interests of their authors.
Therefore, by referencing each other's publications, authors create communities of shared interests.
To enable the study of these communities, we provide the results obtained from the Leiden community detection algorithm~\cite{traag2019louvain} as auxiliary outputs for papers in PubGraph.

\begin{figure}[t!]
  \centering
  \includegraphics[width=0.75\textwidth]{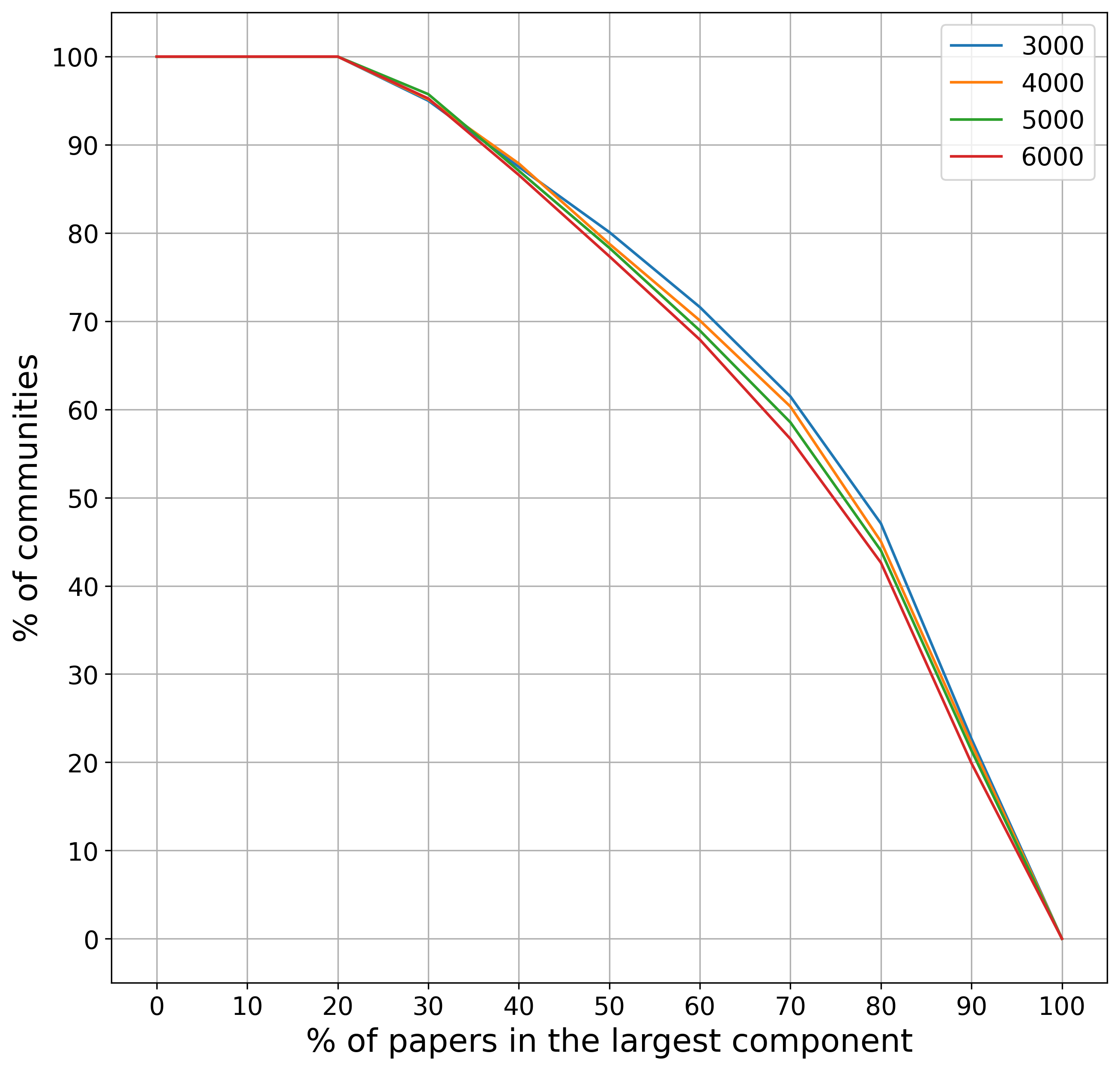}
  \caption{Analysis of the effect of the number of communities on the quality of communities. A higher area under the curve (AUC) indicates more pure communities.}
  \label{fig:app:comm-size}
\end{figure}

To this end, we first extract the full citation network from all the publication-publication links.
Then, we tune the Leiden algorithm\footnote{https://github.com/vtraag/leidenalg} on the extracted citation network with the following parameters: quality function $\in \{ \text{Modular}, \text{RBER}, \allowbreak \text{Significance}, \text{Surprise} \}$, maximum papers per community $\in \{ \text{300k}, \text{500k}\}$, and number of communities $\in \{ 3000, 4000, 5000, 6000\}$.
To evaluate the communities' quality, we use a purity proxy metric extracted from the ancestral graph of the concepts connected to the publications in OpenAlex.
Specifically, we count the number of children for each root concept and select the largest root concept for each community.
Then, we calculate the percentage of the papers that are children of that root concept as the proxy metric.
Figure \ref{fig:app:comm-size} illustrates our results on different numbers of communities.
Based on our experiments, the highest quality communities are produced by the following parameters: quality function $=$ Significance, maximum papers per community $= \text{300k}$, and number of communities $= 3000$.

\subsubsection{Large Language Models}
PubGraph was developed to enable researchers to study scholarly works from a graph perspective.
Through PubGraph, it is possible to learn representations for papers using graph-based methods, which then could be used for various downstream tasks.
Orthogonal to this relational and structural information, are textual information based on scholarly works' content.
When available, textual features complement the graph-based features and can improve the performance of the models \cite{berrebbi2022graphcite}.

Recently, many large language models (LLM) have been introduced to tackle the problem of generating representations for scientific documents~\cite{beltagy-etal-2019-scibert,cohan-etal-2020-specter}.
These pre-trained models are specifically tuned for scientific data and could be used to generate low-dimensional embeddings for input documents.
In this work, to further enable multi-view studies of PubGraph, we provide embeddings generated by LLMs for all the papers.
These embeddings also save resources for researchers who want to use textual information.
To this end, first, we obtain a representing text by concatenating the title and the abstract of each work.
This approach allows us to cover all the works with at least one of these attributes available, improving the general coverage of this data.
Then, we run the representing text through the SciNCL model~\cite{ostendorff-etal-2022-neighborhood} to obtain the embeddings, with each generated embedding being a 768-dimensional vector.
All the generated embeddings are released with an index to match the corresponding papers.



\section{Knowledge Graph Completion}

\begin{figure*}[t]
  \centering
  \includegraphics[width=\textwidth]{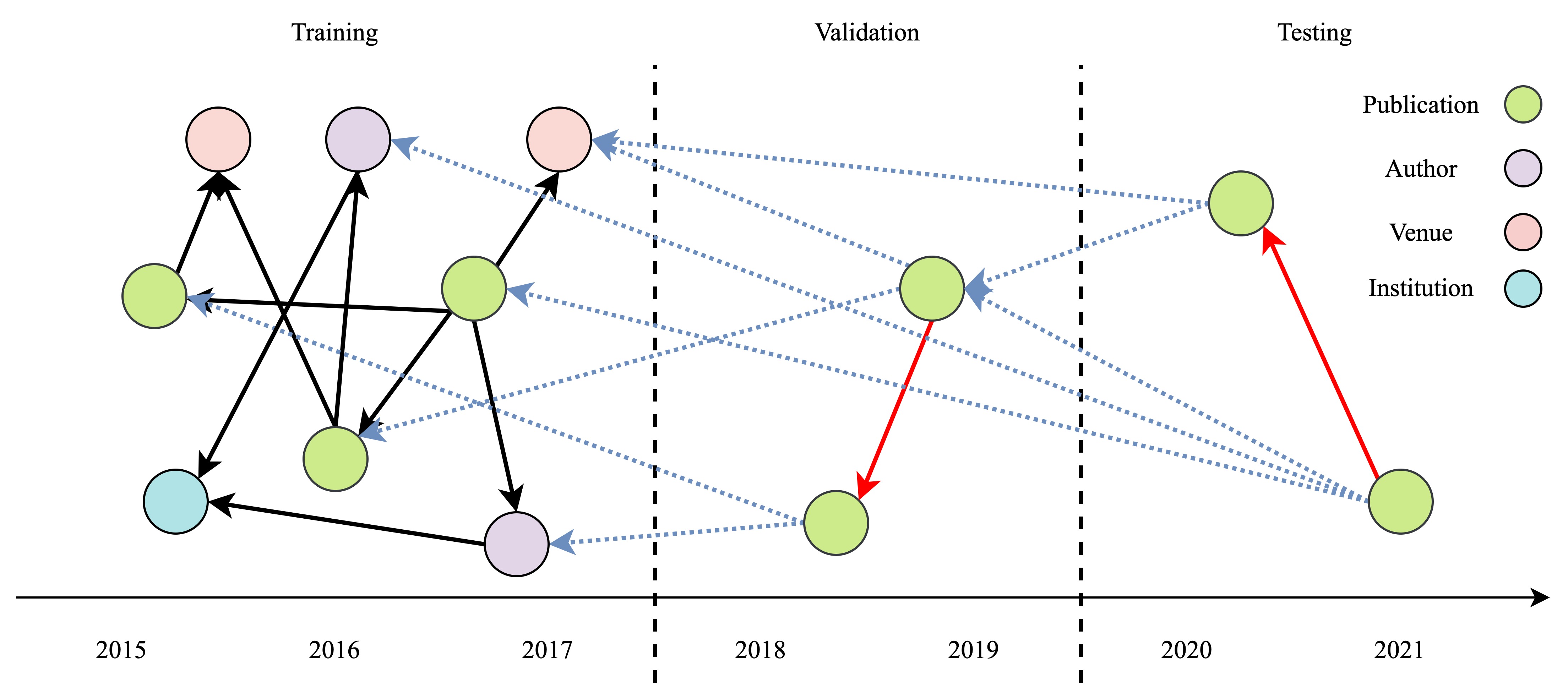}
  \caption{Overview of the training and evaluation scheme. Intra-period current links (black) are used for training in all experiment settings. Intra-period future links (red) are used for evaluation in both validation and testing phases in all experiment settings. Exo-period links (dotted blue) are used in the training phase in transductive settings; however, in inductive settings, these links are only used as auxiliary links during the evaluation phase. Auxiliary links establish connections between seen training nodes and unseen evaluation nodes.}
  \label{fig:intro:over}
\end{figure*}

Traditionally, knowledge graph embedding (KGE) models~\cite{sun2019rotate,trouillon2016complex} have been evaluated in an interpolated, transductive KGC setting where all entities, e.g., papers and authors, are known.
However, one of the challenging aspects of studying scientific progress is dealing with new publications which require inference over unseen samples.
A better-aligned evaluation setting for this purpose is the extrapolated, inductive setting.
An inductive setting requires models to make predictions over previously unseen entities.
While KGs capture the structure necessary for this setting, many models do not address this use case.
Moreover, extrapolated prediction requires train and test sets to be partitioned by a temporal threshold, so model predictions are for a future time epoch.

In this work, we introduce new resources and benchmarks in the extrapolated setting for both inductive and transductive models, framing the research question as a KGC task and supporting the study of this problem from a purely structural standpoint at different scales and across various models.
Moreover, we also introduce a community-based adversarial evaluation setting to 1) mitigate the influence of random negative sampling (due to the scale) in the evaluation phase and 2) maintain the same level of difficulty as evaluated on all of the entities.
Fig. \ref{fig:intro:over} presents an overview of the training and evaluation schemes for the KGC benchmarks in both transductive and inductive settings.
The rest of this section is organized as follows: Sec. \ref{pubg:samp} describes the methodology used to create PG-X benchmarks, Sec. \ref{pubg:qual} presents a data quality analysis over the extracted samples, and Sec. \ref{pubg:adv} presents a set of adversarial evaluation settings for the KGC tasks.

\subsection{Building PG-X Benchmarks}
\label{pubg:samp}

The full PubGraph KG contains a vast amount of information in the form of literal values and sparse properties that are not easily usable by many KG models.
We extract subsets of PubGraph, designated as PG-X, to create easier-to-use benchmarks for KG models.
To extract PGs from the transformed data, we first remove all the publications with no citations that do not cite any other papers to get PG-Full.
Since these nodes are disconnected from other publications, this step mitigates the sparsity problem and reduces the KG size by a large margin.

Given the enormous size of the PG-Full, we create two small and medium-sized sub-KGs to allow future studies at different scales.
To this end, we use snowball sampling~\cite{goodman1961snowball} to extract PG-1M and PG-10M with 1M and 10M publication nodes, respectively.
After sampling, we remove any publication without a publication date.
Next, we extract all the ``cites work (P2860),'' ``author (P50),'' ``published in (P1433)," and ``affiliation (P1416)" links for the sampled publications.
We ensure to include all the available author, source, and institution links from the sampled publications in the benchmarks.
Finally, we split all the benchmarks temporally, using all the publications before 2017 for training, 2017 up until 2020 for validation, and 2020 onward for testing.
Table \ref{tab:aux:data:split} presents the statistics on the extracted splits of each benchmark.

\subsection{Data Quality}
\label{pubg:qual}

\begin{table}[t]
    \caption{Statistics of PG-X benchmarks splits.}
    \label{tab:aux:data:split}
    \centering
    \begin{tabular}{l|cccc}
        \toprule
        Benchmark & \makecell{\#Training\\(Validation)} & \makecell{\#Training\\(Testing)} & \#Validation & \#Test \\ \midrule
        \midrule
        PG-1M & 18.2M & 20.5M & 265k & 146k \\ \midrule
        PG-10M & 269.0M & 305.9M & 3.1M & 2.3M \\ \midrule
        PG-Full & 1.88B & 2.17B & 28.1M & 26.3M \\ \midrule
        \bottomrule
    \end{tabular}
\end{table}

\begin{table}[t]
    \caption{Validity and completeness metrics of sampled KGs.}
    \label{tab:data:qual}
    \centering
        \begin{tabular}{l|ccc}
        \toprule
        Metric     & PG-1M  & PG-10M & PG-Full \\ \midrule
        \midrule
        \makecell[l]{Mutual Citations}           & 0.03\% & 0.04\% & 0.06\% \\
        \makecell[l]{Authorship Completeness}    & 99.97\% & 99.97\% & 99.92\% \\
        \makecell[l]{Venue Completeness}         & 92.37\% & 90.25\% & 75.34\% \\
        \makecell[l]{Institution Completeness}   & 81.45\% & 71.21\% & 45.77\% \\ \midrule
        \bottomrule
    \end{tabular}
\end{table}

To evaluate the quality of the extracted benchmarks, we check the validity and completeness of our KGs.
For validity, we look for potential mutual citations, cases where two papers reference each other, violating strict temporal order. This artifact may appear when articles have several revisions, but OpenAlex only reports the earliest publication date.
For completeness, we calculate publication-author, publication-source, and author-institution relations completeness.
Table \ref{tab:data:qual} showcases these metrics on the extracted KGs.
As evident from the metrics, all the benchmarks exhibit an extremely low mutual citations percentage which is evidence of their quality.
Moreover, the small and medium-sized KGs exhibit higher completeness metrics which we attribute to the forced inclusion of all authors, venues, and institutions links.

\subsection{Adversarial Evaluation Setting}
\label{pubg:adv}

\begin{table}[t]
    \caption{Negative sampling results on the PG-1M benchmark.}
    \label{tab:exp:negative}
    \centering
    \begin{tabular}{l|c|cccc}
        \toprule
        Variation & \makecell{\#Negative Samples} & MRR & Hits@1 & Hits@10 & \makecell{Time (Seconds)} \\ \midrule
        \midrule
        Random & 1000 & 0.723 & 0.608 & 0.918 & 588 (CPU) \\ \midrule
        Entity Type & 1000 & 0.560 & 0.418 & 0.826 & 655 (CPU) \\
        \makecell{Time Constrained} & 1000 & 0.577 & 0.449 & 0.817 & 601 (CPU) \\
        Community & 1000 & 0.076 & 0.023 & 0.167 & 1008 (CPU) \\
        Full & $\sim$3.38M & 0.015 & 0.000 &  0.036 & 81987 (GPU) \\ \midrule
        \bottomrule
    \end{tabular}
\end{table}

\begin{figure}[t!]
  \centering
  \includegraphics[width=0.7\columnwidth]{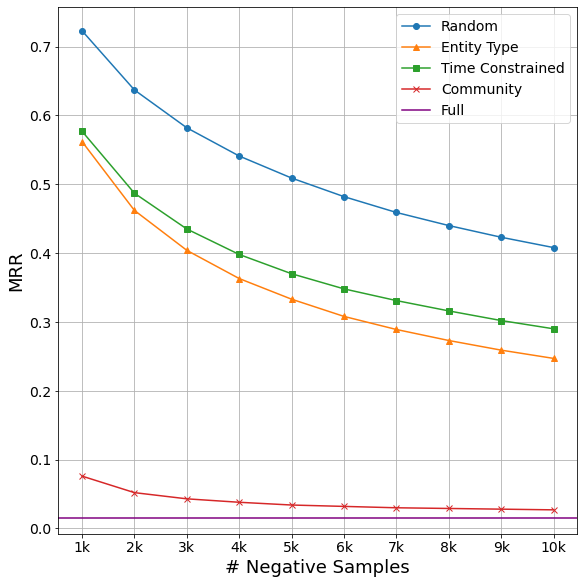}
  \caption{Analysis of the effect of negative samples count on the model's performance measured by MRR.}
  \label{fig:app:negative}
\end{figure}

One of the most common strategies to evaluate KGC on large-scale graphs is to sample a fixed number of negative samples for each positive sample during the evaluation phase.
However, this strategy is prone to exhibiting inflated performance due to having no control over the difficulty of the sampled nodes.
Moreover, calculating the evaluation metrics on the complete set of samples becomes increasingly more expensive as the size of the KG grows.
Hence, we propose three alternative strategies for negative sampling during the evaluation phase.
These strategies aim to find an efficient method to be used as a proxy for complete metric calculations.
Our proposed strategies are as follows:
\begin{enumerate}
  \item \textbf{Entity Type}: This is the most straightforward strategy in which we only sample candidate nodes with the same type as the target node.
  For example, in our case, we only sample from the publications.
  \item \textbf{Time Constrained}: Building upon our first strategy, we further add the constraint of only sampling candidate nodes from the nodes within the evaluation period.
  Intuitively, these unseen (inductive) or less seen (transductive) nodes will pose more problems for the model during the evaluation phase.
  \item \textbf{Community}: Given a target node, we sample candidate nodes only from its community.
  This strategy relies on the auxiliary outputs, i.e., communities, generated as described in Sec. \ref{pubg:aux}.
  We hypothesize that these nodes pose the most difficulty for the model during the evaluation phase.
\end{enumerate}

To test the proposed strategies, we train a ComplEx~\cite{trouillon2016complex} model using the DGL-KE toolkit~\cite{zheng2020dgl}.
We tune the hyper-parameters of our model using the following set of values:
\textit{embedding dimensions} $\in \{ 50, 100, 200, 400 \}$,
\textit{learning rate} $\in\allowbreak \{ 0.003\allowbreak, 0.01\allowbreak, 0.03, 0.1, 0.3 \}$,
\textit{number of negative samples} $\in\allowbreak \{ 128\allowbreak, 256, 512\allowbreak, 1024, 2048 \}$, and
\textit{regularization coefficient} $\in \{ 0.0, \text{1e-9}, \text{1e-8}, \text{1e-7}, \text{1e-6}, \text{1e-5} \}$.
Table \ref{tab:exp:negative} presents the results of our experiments with the aforementioned negative sampling strategies in the evaluation phase.
The reported times are for one evaluation run over the complete testing set of the PG-1M benchmark ($\sim$147K samples).
As evident from these results, the community-based method is the best proxy to the full metrics calculation while still being significantly time efficient.
Even if we factor in the 11.5 hours (41400 seconds) that it takes to learn communities for all the 91M publications, the difference in computation time becomes much more significant when we have to repeat the evaluation process over and over again, e.g., for validation, fine-tuning, etc.
Moreover, the full metrics are calculated on a GPU which is far more efficient than the calculations on the CPU.
It is important to note that the community-based method is helpful in evaluation settings where the ground truth is known; however, in settings where the ground truth is unknown, e.g., a deployed model, there is no workaround to complete ranking computations as we have to consider all the entities regardless.

We further analyze the effect of the number of negative samples on the model's performance.
Figure \ref{fig:app:negative} presents the result of our experiments with varying numbers of negative samples on all the introduced strategies.
As expected, the model's performance rapidly drops with the increase of negative samples.
Moreover, the community-based negative sampling results act as an excellent proxy at 5k negative samples and seem to converge to the full variation around 10k negative samples.
This finding is further evidence of the effectiveness of this method.

%% file: sections/related.tex
\section{Related Works}


\subsection{Scientific Knowledge Graphs}
In recent years, a wide range of scientific KGs (SKG) have emerged in the research community.
Examples of these SKGs are Scholia~\cite{NielsenF2017Scholia}, ORKG~\cite{stocker2023fair}, OpenAIRE~\cite{manghi_paolo_2019_2643199}, and MAG240M~\cite{hu2021ogb}.
Each of the aforementioned SKGs has different characteristics that make them unique and interesting to the community.
Table \ref{tab:rel:skg} compares PubGraph with the existing SKGs across various properties.
As evident from this table, PubGraph is built on a more grounded ontology and provides much more information and artifacts compared to other SKGs.

\begin{table}[t!]
    \caption{Comparison between PubGraph and the existing SKGs.}
    \label{tab:rel:skg}
    \centering
    \resizebox{\columnwidth}{!}{
    \begin{tabular}{l|cccccc}
        \toprule
        SKG & \#Articles & Source & Ontology & Embeddings & Community & \makecell{External Links\\(Other Sources)} \\ \midrule
        \midrule
        Scholia & 39M & Wikidata & Wikidata & \xmark & \xmark & \xmark \\ \midrule
        ORKG & 25k & Curated & Proprietary & \xmark & \xmark & \xmark \\ \midrule
        OpenAIRE & 164M & Curated & Proprietary & \xmark & \xmark & \xmark \\ \midrule
        MAG240M & 121M & MAG & Proprietary & \cmark & \xmark & \xmark \\ \midrule
        PubGraph & 250M & OpenAlex & Wikidata & \cmark & \cmark & \cmark \\ \midrule
        \bottomrule
    \end{tabular}
    }
\end{table}

\begin{table}[t]
    \caption{Statistics of extracted benchmarks compared to the existing large-scale KGC benchmarks. As evident, PG-Full has more than 2x nodes and 3.6x edges compared to the largest existing benchmarks.}
    \label{tab:data:stats}
    \centering
    \begin{tabular}{l|rrr}
        \toprule
        Benchmark     & \#Nodes  & \#Edges  & \#Relations \\ \midrule
        \midrule
        ogbl-citation2~\cite{hu2020open} & 2,927,963 & 30,561,187 & 1 \\
        Freebase~\cite{bollacker2008freebase} & 86,054,151 & 338,586,276 & 14,824 \\
        WikiKG90Mv2~\cite{hu2021ogb} & 91,230,610 & 601,062,811 & 1,315 \\ \midrule
        PG-1M       & 3,378,202 & 22,442,976 & 4 \\
        PG-10M      & 25,312,490 & 315,225,337 & 4 \\
        PG-Full     & 184,126,885 & 2,201,239,147 & 4 \\ \midrule
        \bottomrule
    \end{tabular}
\end{table}

\subsection{Large Scale KGC Benchmarks}

KGC is one of the most common tasks defined on KGs.
Recent efforts~\cite{hu2021ogb,hu2020open} have shifted toward introducing more large-scale benchmarks for KGC; however, there is still a shortage of benchmarks for large-scale graph learning.
We believe the PG-X benchmarks introduced in this paper can help mitigate this shortage.
Table \ref{tab:data:stats} showcases the statistics of the sampled KGs along with a comparison to existing large-scale KGC benchmarks in the literature.
As evident from the numbers, PG-X benchmarks provide an opportunity to evaluate KG models on larger (2x nodes and 3.6x edges) and more flexible (3.3M to 184M range) benchmarks.

%% file: sections/conclusion.tex
\section{Conclusion and Future Work}


In this work, we introduced PubGraph, a new large-scale resource in the form of a KG built on Wikidata ontology and extracted from the OpenAlex catalog with more than 13B edges and 385M nodes.
As presented through different comparisons, PubGraph provides a much-needed unifying and comprehensive resource for researchers to study scientific progress that connects multiple sources.
PubGraph also enables the study of scientific documents from distinct perspectives through the information extracted from auxiliary community detection algorithms and large language models.
Moreover, we created three KGC benchmarks with varying sizes to enable future studies at different scales and for both transductive and inductive settings.
Finally, we identified a set of challenging adversarial evaluation settings for the introduced  benchmarks that overcome the common downfall of large-scale KGC evaluation settings.
As for future directions for PubGraph, one direction is to improve the coverage of connections to external sources.
Moreover, it is possible to bring in more external data sources, e.g., SKGs such as Scholia, and link them with PubGraph.
Finally, another venue is to add other metadata that is of interest to the community, such as awards and grants, which further enables researchers to study these events in the larger context.